\documentclass[conference]{IEEEtran}
\IEEEoverridecommandlockouts
\usepackage{cite}
\usepackage{amsmath,amssymb,amsfonts}
\usepackage{algorithmic}
\usepackage{graphicx}
\usepackage{textcomp}
\usepackage{comment}
\usepackage{kotex}
\usepackage{xcolor}
\usepackage{xcolor,colortbl}
\usepackage{subfigure}
\usepackage{makecell}
\usepackage{mathtools}
\usepackage{textcomp}
\usepackage{setspace}
\usepackage{pifont}
\usepackage{soul}
\usepackage{siunitx}
\usepackage{wrapfig}
\usepackage[normalem]{ulem}
\usepackage{amsfonts}
\usepackage{microtype}
\usepackage{textcase}
\usepackage{multirow}
\usepackage{caption}
\usepackage{subcaption}
\usepackage[colorlinks=true,linkcolor=purple,citecolor=purple,urlcolor=black]{hyperref}

\newcolumntype{L}{>{\raggedright\arraybackslash}X}

\usepackage{textcomp}
\usepackage{tcolorbox}

\usepackage[linesnumbered,ruled,vlined]{algorithm2e}

\usepackage{titlecaps}
\usepackage{url,tikz}

\newcommand{\bcircled}[1]{\tikz[baseline=(char.base)]{ \node[shape=circle,fill,inner sep=1pt] (char) {\textcolor{white}{#1}};}}

\newcommand{\squishlist}{
\begin{list}{$\bullet$}
	{ \setlength{\itemsep}{0pt}      \setlength{\parsep}{-0pt}
		\setlength{\topsep}{4pt}       \setlength{\partopsep}{0pt}
		\setlength{\listparindent}{-2pt}
		\setlength{\itemindent}{-5pt}
		\setlength{\leftmargin}{1em} \setlength{\labelwidth}{0em}
		\setlength{\labelsep}{0.5em} } }
  
\newcommand{\squishend}{
\end{list}  }
\def\BibTeX{{\rm B\kern-.05em{\sc i\kern-.025em b}\kern-.08em
    T\kern-.1667em\lower.7ex\hbox{E}\kern-.125emX}}
\begin{document}

\title{\huge
Shared Disk KV Cache Management for Efficient Multi-Instance Inference in RAG-Powered LLMs
}

\author{\IEEEauthorblockN{
Hyungwoo Lee$^{1}$, Kihyun Kim$^{1}$, Jinwoo Kim$^{1}$, Jungmin So$^1$, Myung-Hoon Cha$^2$\\ 
Hong-Yeon Kim$^2$, James J. Kim$^3$, Youngjae Kim$^{1,\dagger{}}$
\thanks{$^{\dagger}$Y. Kim is the corresponding author.}
}
\IEEEauthorblockA{$^1$Dept. of Computer Science and Engineering, Sogang University, Seoul, Republic of Korea\\
$^2$ETRI, Daejeon, Republic of Korea,
$^3$Soteria Inc.} 
}

\maketitle
\thispagestyle{plain}
\pagestyle{plain}

\begin{abstract}

Recent large language models (LLMs) face increasing inference latency as input context length and model size continue to grow. In particular, the retrieval-augmented generation (RAG) technique, which enhances LLM responses by incorporating external knowledge, exacerbates this issue by significantly increasing the number of input tokens. This expansion in token length leads to a substantial rise in computational overhead, particularly during the prefill stage, resulting in prolonged time-to-first-token (TTFT). To address this issue, this paper proposes a method to reduce TTFT by leveraging a disk-based key-value (KV) cache to lessen the computational burden during the prefill stage. 
We also introduce a disk-based shared KV cache management system, called Shared RAG-DCache, for multi-instance LLM RAG service environments. This system, together with an optimal system configuration, improves both throughput and latency under given resource constraints. 
Shared RAG-DCache exploits the locality of documents related to user queries in RAG, as well as the queueing delay in LLM inference services. It proactively generates and stores disk KV caches for query-related documents and shares them across multiple LLM instances to enhance inference performance. 
In experiments on a single host equipped with 2 GPUs and 1 CPU, Shared RAG-DCache achieved a 15–71\% increase in throughput and up to a 12–65\% reduction in latency, depending on the resource configuration.
\end{abstract}

\section{Introduction}

Large Language Models (LLMs) have demonstrated exceptional performance across various tasks, and their increasing scale continues to deliver progressively more powerful capabilities. Models with billions or even trillions of parameters have significantly advanced the state-of-the-art in natural language processing, enabling remarkable abilities in understanding context, generating coherent text, and generalizing across diverse linguistic scenarios. Nevertheless, despite their extensive capacity, LLMs frequently encounter difficulties when tasked with providing accurate responses involving the most recent or specialized internal corporate data, as such information typically falls outside the scope of their static pre-training datasets. This limitation arises because these models do not inherently possess mechanisms to dynamically integrate or update knowledge post-training, which significantly restricts their applicability in scenarios requiring up-to-date or confidential domain-specific insights.

To address this limitation, Retrieval-Augmented Generation (RAG)\cite{10.5555/3495724.3496517} has gained attention. RAG improves LLM prompts by retrieving external documents related to the user query, thereby increasing the accuracy of responses regarding up-to-date information or domain-specific knowledge\cite{10.5555/3495724.3496517, siriwardhana-etal-2023-improving, chen2023benchmarkinglargelanguagemodels}. However, incorporating external context documents into the prompt significantly increases its length, leading to longer Time-To-First-Token (TTFT)\cite{fu2024lazyllm}  
and reduced throughput. This phenomenon arises primarily from the increased computational complexity during the prefill phase of LLM inference, where the model computes attention scores and generates the corresponding key-value (KV) matrices for all tokens in the expanded prompt. Specifically, the complexity of KV cache generation scales approximately as 
(\(O(L·N^2·D)\))\cite{vaswani2017attention, NEURIPS2022_67d57c32}, where L represents the number of Transformer layers, N denotes the total token length of the input (including both original prompts and added contexts), and D corresponds to the dimensionality of the hidden representations.

During this process, each token's embedding is transformed into query, key, and value vectors, after which self-attention calculations occur between these vectors, producing the attention scores and resulting value vectors. These calculated key-value pairs, which constitute the KV cache, must be computed for every token within the input prompt during the prefill stage, imposing substantial computational overhead. Particularly, as the input length (L) grows due to appended retrieved documents, the self-attention operations require quadratic complexity (\(O(L·N^2·D)\)), substantially escalating the computation demand and slowing inference speed. This increased complexity becomes especially pronounced with larger LLM models, whose parameter sizes further amplify the computational cost.

Nevertheless, we observe an opportunity arising from the existence of query locality, meaning that a subset of external documents tends to be frequently referenced across multiple user queries in RAG systems. We analyzed popular question and answer datasets and found that processing 50\% of the queries requires only between 3.1~31.39\% of the documents.
This result shown that locality exists in documents referenced by user queries across various workloads, indicating that precomputed KV caches for these commonly accessed documents could significantly reduce redundant computations during inference.

The existing studies, RAGCache\cite{Jin2024RAGCacheEK} and TurboRAG\cite{Lu2024TurboRAGAR}, have explored optimizing RAG inference by precomputing KV caches for frequently accessed external documents. 
RAGCache utilizes a multi-level caching approach, leveraging GPU and CPU memories to dynamically cache intermediate KV states, which significantly reduces inference latency. However, it faces limitations due to the restricted capacities of GPU and CPU memory, making it challenging to store large numbers of frequently referenced documents. On the other hand, TurboRAG adopts a disk-based caching strategy, storing precomputed KV caches offline, and significantly decreasing the prefill computational overhead and inference latency. Yet, TurboRAG does not explicitly address the multi-instance or multi-host scenarios for KV cache sharing, which are essential for scaling real-world RAG services.

In contrast, our approach clearly differentiates itself by emphasizing disk-based KV cache storage to effectively manage the growing size of KV caches caused by larger LLM parameters and longer input prompts in RAG scenarios. By utilizing persistent disk storage, our method overcomes the capacity constraints associated with GPU and CPU memory, ensuring effective and semi-permanent storage. Furthermore, we explicitly support multi-instance and multi-host LLM service environments, facilitating efficient sharing and reuse of KV caches stored on disk. This design leverages the infrequent update nature of external documents, allowing persistent, shareable KV caching across multiple inference instances and hosts—capabilities not fully addressed by existing methods.


To leverage this opportunity, we propose a disk-based KV cache management system composed of two solutions: \textit{RAG-DCache} and \textit{Shared RAG-DCache}. RAG-DCache precomputes and stores the KV cache for frequently retrieved document chunks within a disk-resident vector database. During inference, these precomputed KV caches are reused directly, eliminating the costly recomputation of the full document context. Shared RAG-DCache extends this concept to multi-instance inference environments, enabling multiple LLM instances to share a common KV cache stored on disk, thus further enhancing inference performance by proactively generating and distributing KV caches across instances during query waiting periods. 

The proposed system is consist of three key components:
\squishlist
\item
\textbf{KV Cache Manager}: Responsible for the offline precomputation and management of KV caches for document chunks, maintaining them in disk-based storage integrated within the vector database.
\item
\textbf{KV Cache Generator}: Operates proactively, especially in multi-instance settings, to prefetch and generate KV caches during query wait times, thus efficiently utilizing idle resources.
\item
\textbf{Prompt Generator (RAG Processor)}: Combines the retrieved KV caches with user queries to construct the final inference prompts, enabling the LLM to bypass redundant computations by directly leveraging pre-stored KV caches.
\squishend

To demonstrate the effectiveness of our proposed system, we conducted experiments on a server equipped with dual GPUs and a single CPU, using the SQuAD\cite{rajpurkar2016squad} dataset as representative workload. 
Results showed that employing RAG-DCache reduced the TTFT by approximately 10\%–20\%, with throughput increasing significantly as the model size and batch size grew. 
Furthermore, our multi-instance solution, Shared RAG-DCache, achieved even more substantial improvements, increasing throughput by up to 71\% and reducing latency by up to 65\%.
\section{Background and Motivation}
\subsection{KV Cache Utilization in LLM Inference}
Transformer-based LLMs\cite{vaswani2017attention} generate text in an autoregressive manner by producing one token at a time. To generate each token, the model processes the entire prompt and then, during the decode phase, reuses the previously generated tokens as input to predict the next token\cite{10.1145/3600006.3613165, Wu2023FastDI, xiao2024efficient, 280922, 298687}. Recomputing all tokens at every generation step would be inefficient, so the model stores the Key and Value matrices from previous steps in GPU memory as a cache, which is then reused for subsequent token predictions. 
This KV cache is critical for avoiding redundant computations and helps reduce the overall complexity of the decode phase. For example, optimization libraries like DeepSpeed-Inference\cite{rasley2022deepspeed} incorporate KV caching to enhance inference efficiency in large Transformer models. 

\subsection{RAG Prompt Composition}
In a RAG system, external documents are retrieved using a vector database. These documents are first divided into manageable chunks, then converted into vector embeddings using an embedding model. These embeddings, together with their document IDs and original texts, are stored in the vector database\cite{karpukhin-etal-2020-dense, 8733051}. When a query is received, it is similarly embedded and matched against this vector store to retrieve the top-k most relevant document chunks. The final LLM prompt is then composed by concatenating the retrieved document texts with the user query, typically structured as: "Document: (retrieved texts) + Query: (user question) + Answer:". However, as the number of tokens in the prompt lengthens due to added context, the computational complexity during the prefill phase increases significantly, leading to higher TTFT and reduced throughput.

To address this issue, this paper proposes utilizing external knowledge documents not only as retrieval sources but also as precomputed Key-Value (KV) caches. Specifically, if the KV tensors of the external knowledge documents are precomputed and stored, they can be directly reused whenever the same document is included in future LLM prompts. By leveraging disk-based caching of these precomputed KV tensors, the costly prefill computations associated with document processing are significantly reduced, resulting in shorter TTFT and improved inference performance. However, the effectiveness of this caching mechanism fundamentally depends on query locality—the frequency with which particular documents are repeatedly referenced across multiple queries. Hence, understanding and exploiting query locality becomes crucial for optimizing system performance, which we discuss further in the following section.


\subsection{Locality of Documents Retrieved by Queries in RAG}

To verify the locality between queries and their retrieved documents when using RAG, we used popular question-answering datasets, HotpotQA\cite{yang2018hotpotqa}, SQuAD\cite{rajpurkar2016squad} and TriviaQA\cite{joshi2017triviaqa}. We first constructed a FAISS vector database by embedding documents(using all-MiniLM-L6-v2\cite{wang2020minilmv2} embedding model) from each dataset. Then, for each query, we measured the similarity between the query embeddings and the document embeddings using an inner-product-based similarity metric, retrieving the top-1 most similar document to query. And using the [query, top-1 document] data, we calculated the proportion of documents required to process the queries.

\begin{figure}[htbp]
\centerline{\includegraphics[width=\linewidth]{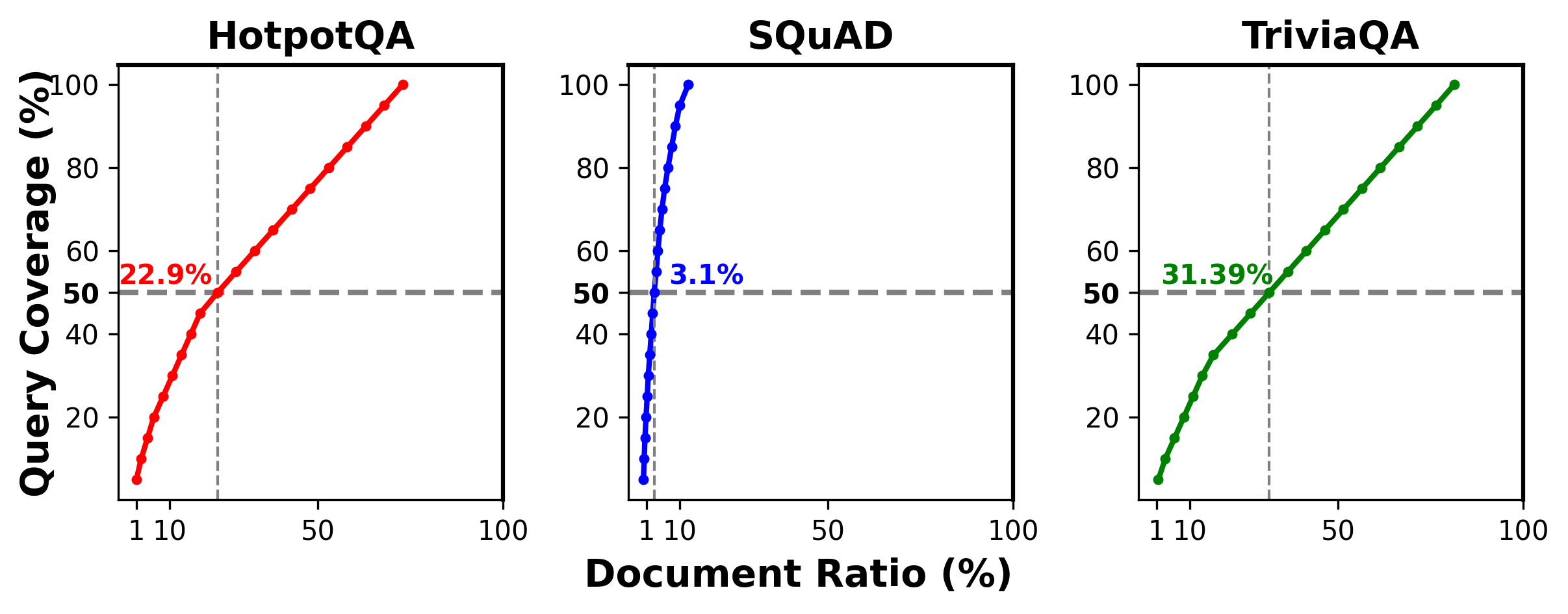}}
\caption{CDF of query related documents.}
\label{fig:back_cdf}
\end{figure}

The results(Figure~\ref{fig:back_cdf}) show that only 22.9\%, 3.1\%, and 31.4\% of the most frequently retrieved documents account for 50\% of all queries in each dataset. This result suggests that caching a small subset of frequently accessed documents can effectively serve a large portion of queries, highlighting the efficiency of using document caching in RAG systems.


\subsection{Shared KV Cache for Multi-instance LLM Inference}


LLM-based inference services generally operate multiple model instances in parallel to handle numerous real-time user requests. 
For example, on a server with two GPUs, two LLM instances can be run to process two queries simultaneously, or one GPU can be allocated to a different task. In such multi-instance environments, each instance performs inference independently, so an instance cannot inherently access the KV cache computed by another instance. Therefore, to maximize the benefits of caching, a structure is needed that allows instances to exchange cache data via shared memory or storage (e.g., disk). We focus on a disk-based KV cache sharing approach to cope with the increasing number of LLM parameters and the growing size of input tokens.

Moreover, as requests(queries) per second increase with rising service loads, requests that exceed the capacity of an individual instance incur queue wait time. In a single-instance LLM environment, only one request can be processed at a time, so subsequent requests must inevitably wait. However, in a multi-instance environment, there is a possibility of utilizing free resources or other devices to prepare tasks that are waiting in the queue. For instance, while one GPU is decoding a current query, it may be possible to leverage another idle GPU or a CPU to carry out preliminary work for the next query, thereby reducing response latency. The proposed Shared RAG-DCache implements this idea by prefetching necessary document KV caches for requests that wait in the inference service queue beyond a certain threshold. As a result, when the request eventually reaches an LLM instance, it can focus solely on the decoding step.

\begin{figure}[htbp]
\centerline{\includegraphics[width=\linewidth]{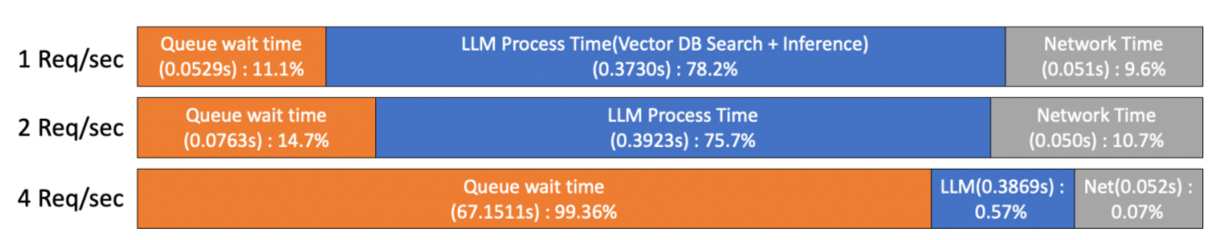}}
\caption{Inference Latency Based on Requests Per Second in a Multi-Instance LLM(with RAG) Service Environment.}
\label{fig:back_lat}
\end{figure}

Figure~\ref{fig:back_lat} shows the average response time from when a client issues a query until the response is received—divided into queue wait time, LLM processing time, and network time(communication time between client and LLM server)—under varying query rates per second, using a single Llama-3.2-1B\cite{Touvron2023LLaMAOA} model on one GPU. For detailed experimental settings, refer to Table~\ref{tab:tab0} in Section 4.
At 1–2 queries per second, LLM processing time makes up over 70\% of the total time. However, once the query arrival rate exceeds the LLM processing time, the queue wait time increases exponentially. This suggests that under heavy loads, it is beneficial to utilize the waiting period for prefetch operations. Therefore, in a multi-instance environment, it is desirable to share caches among multiple instances to avoid redundant computation, and prepare caches in advance during wait times, allowing the GPU to focus on pure inference tasks. Shared RAG-DCache is the system designed to meet these needs.

\section{Design and Implementation}

In this section, we detail the architecture and operation of our proposed RAG-DCache system—a disk-based KV cache for single-instance LLM inference—as well as its extension to a multi-instance environment, called Shared RAG-DCache.

\subsection{Disk-based KV Cache Structure and Operation}
The RAG-DCache system extends the traditional Retrieval-Augmented Generation pipeline by adding a disk-resident Key–Value cache storage and associated management modules. The basic idea is to precompute the KV cache for each document chunk in the vector database and store these caches on disk so that they can be reloaded during inference instead of recomputed from the original text. 

\begin{figure}[htbp]
    \centering
    \subfigure[General vector DB]{
        \includegraphics[width=1\linewidth]{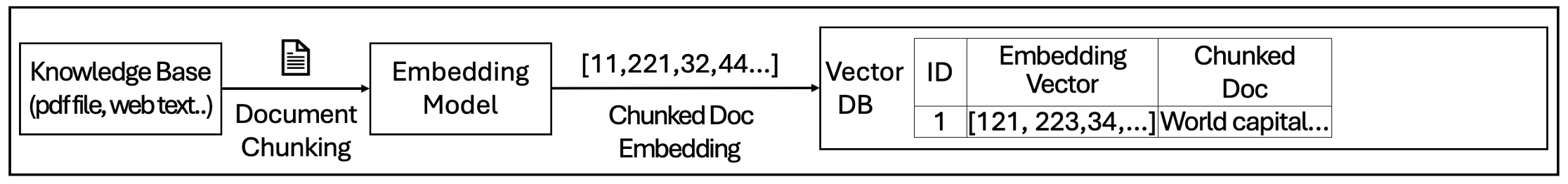}
        \label{fig:des_vdb1}
    }
    \hfill
    \subfigure[Vector DB with additional utilization of KV Cache]{
        \includegraphics[width=1\linewidth]{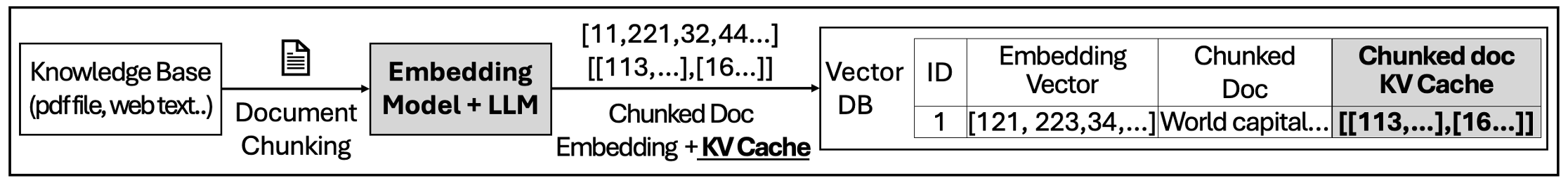}
        \label{fig:des_vdb2}
    }
    \caption{Vector DB creation process and tuple structure for RAG.}
    \label{fig:des_vdb}
\end{figure}

Figure~\ref{fig:des_vdb} contrasts a standard vector database with our augmented version. In a conventional RAG setup, the vector database stores embeddings of document chunks for similarity search. 
In our approach RAG-DCache, we augment each stored document chunk with its precomputed “chunked-document KV cache,” pairing it with the document’s ID and embedding in the database. 
Because document data changes rarely once the vector DB is built, we can leverage idle hardware to generate these KV caches offline and persist them to disk. (It is also possible to generate a KV cache on-the-fly during inference for a newly encountered document and then add it to the DB for future reuse.) 

By caching each document’s transformer key-value pairs in advance, the LLM can skip directly to using this cached representation when that document is retrieved for a query, rather than processing the raw text each time.

\begin{figure}[htbp]
    \centering
    \includegraphics[width=1\linewidth]{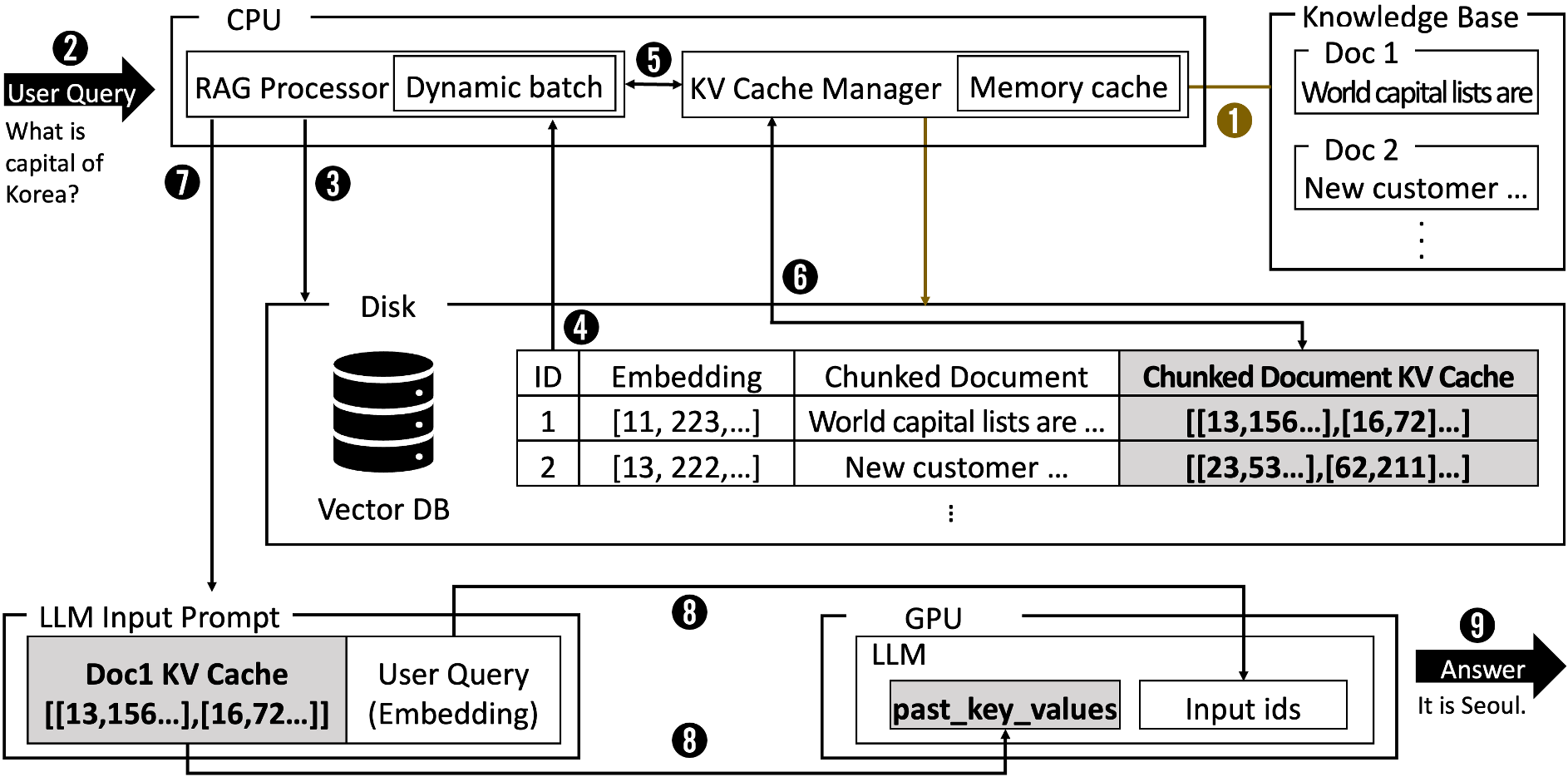}
    \caption{Design and Operation of RAG-DCache.}
    \label{fig:des_stov}
\end{figure}

Figure~\ref{fig:des_stov} illustrates the design and operation of RAG-DCache, which incorporates a KV Cache–linked vector DB for LLM inference using RAG. The main Components of disk-based KV Cache are as follows:

\squishlist
\item 
\textbf{KV Cache Manager}: This module is responsible for creating and managing the stored caches. It generates the KV cache for document chunks using the LLM and stores the resulting key-value tensors in the vector database. It also handles retrieval of these caches from disk upon request. 
To minimize disk I/O latency, the KV Cache Manager employs an in-memory(CPU RAM) cache to hold frequently or recently used KV entries, leveraging faster memory access and reducing repeated disk reads. In essence, it acts as the interface between slow persistent storage and the rest of the system, optimizing cache generation and lookup.

\item 
\textbf{RAG Processor}: The RAG Processor orchestrates the RAG inference workflow. Upon receiving a user query, it performs similarity search on the vector database to fetch the 
relevant document ID. It then requests the corresponding KV caches for the document from the KV Cache Manager, and composes the final LLM prompt by combining the query with the retrieved KV caches. 

\item 
\textbf{Integrated Vector Database}: This is an extended vector store that holds not only each document chunk’s embedding and original text, but also the precomputed KV caches. 
Each entry in the vector DB effectively becomes a tuple of the form (embedding, document ID, text, KV cache). The inclusion of the KV cache alongside the embedding means that after retrieval, the system immediately has access to the document’s encoded representation for the LLM. Since documents are largely static, these caches can be generated offline and remain valid unless the document content changes. For any new documents added to the corpus, on-demand cache generation can be performed and the caches appended to the database, keeping the cache store up-to-date. This integrated DB design ensures that the vector index serves a dual purpose: it provides nearest-neighbor search for relevant documents and acts as a lookup table for their cached LLM representations.
\squishend

The end-to-end operation of RAG-DCache proceeds as follows (refer to the numbered steps in Figure~\ref{fig:des_stov}). 

\bcircled{1} Offline Cache Preparation: Initially, use existing documents to pre-generate the KV caches and build a KV-augmented vector database. The KV Cache Manager takes each document (or document chunk) and computes its KV cache using the LLM model, then stores this cache in the vector DB alongside the document’s embedding and ID. This step can be done offline or in the background, populating the disk cache before queries arrive. By the end of this step, the system has a disk-based cache of key-value pairs ready for many documents in the corpus.

\bcircled{2}, \bcircled{3} Query Retrieval: When a user query comes in, the RAG Processor embeds the query and searches the vector database for the most relevant document ID. This yields the ID of the document that will be used to augment the query.


\bcircled{5}, \bcircled{6} KV Cache Retrieval: For each document ID obtained in step \bcircled{2}, \bcircled{3}, the RAG Processor requests the corresponding KV cache from the KV Cache Manager. The KV Cache Manager checks its memory cache for the entry; if present, it returns it immediately from RAM. If not, it loads the KV cache from disk storage into memory and returns it to the RAG Processor.

\bcircled{7}, \bcircled{8} Prompt Composition: Meanwhile, the user’s query text is converted into the appropriate embedding or token IDs for the LLM model if not already done during retrieval. Once the KV cache for the document is in hand, the RAG Processor constructs the final LLM input prompt. It does this by inserting the retrieved KV cache data into the model’s context as “past key-values” and appending the user query tokens as the current input. 
Essentially, the LLM is tricked into believing it has already processed the retrieved documents, because their resulting key-value pairs are provided, and now it only needs to attend to the user’s query. In practice, this means setting the model’s internal key-value state to the cached values and providing the query tokens as the next sequence to process.

\bcircled{9} LLM Inference (Prefill + Decode): The LLM, now armed with the combined prompt (document KV Cache + query embeddings), 
proceeds to generate an answer. It first goes through its prefill phase and then the decode phase to produce output tokens. Because of RAG-DCache, the prefill phase is dramatically accelerated: instead of recomputing key-value pairs for the document’s text on the GPU, the model directly uses the precomputed keys and values. It only needs to encode the user query itself and then can immediately attend to the cached document representations when predicting the answer. 
After the prefill, the decode stage proceeds as usual to generate the response token by token. 


By leveraging pre-stored caches in this manner, RAG-DCache reduces the TTFT and overall computational load on the GPUs. In a baseline RAG system without caching, the LLM must process the full text of the retrieved documents for every single query, leading to significant repeated work in the prefill stage. 
Our approach avoids this repetition. There is an overhead for loading the KV cache from disk (when a cache is not already in memory), but as long as efficient storage (fast SSD) and caching strategies are used, the sum of “(disk load time) + (cached prefill time)” is typically much less than the original prefill time required to encode the documents from scratch. 
In other words, even accounting for disk I/O, the TTFT with RAG-DCache is lower than without it, provided the caches are effectively utilized. This will be quantitatively demonstrated in our evaluation. Note that when multiple documents are retrieved (k \textgreater 1), we do not calculate cross-attention between the documents we only calculate KV values between the user query and the retrieved documents. This may lead to accuracy degradation as shown in the evaluation(Figure~\ref{fig:exp_accu})

To address this issue, instead of precomputing each document’s KV Caches and storing them in the vector DB, we changed our approach so that during inference—when RAG retrieves documents—the KV Caches for the top-k documents are generated and stored together with the vector DB. We also made sure that the generated KV Caches are stored and managed within the vector DB along with the documents’ IDs, allowing us to handle any top-k scenario. For example, if top-k = 3 and the retrieved document IDs are 1, 2, and 3, we calculate the attention for those three documents together to generate their KV Caches, then add that combination of document IDs to the vector DB so the KV Cache Manager can easily locate them. This approach applied to Shared RAG-DCache, which will be described next, leverages idle time during the inference process to precompute KV caches, regardless of the number of retrieved documents. These precomputed KV caches are then stored on disk, shared across instances, and reused to improve efficiency and maintain accuracy.


\subsection{Multi-Instance Structure of RAG-DCache}
While RAG-DCache improves single-instance performance, modern LLM services often deploy multiple LLM instances (across one or more GPUs or nodes) to handle high query throughput. In such environments, caches computed in one instance could be beneficial to others. We therefore introduce Shared RAG-DCache, an extension of RAG-DCache for multi-instance LLM service environments. Shared RAG-DCache enables cache sharing and cache prefetching across multiple parallel LLM inference processes. 
The goal is to exploit both the locality of document usage across different queries and the idle time that queries spend waiting in a queue due to high load to proactively generate and distribute KV caches. 

\begin{figure}[htbp]
    \centering
    \includegraphics[width=1\linewidth]{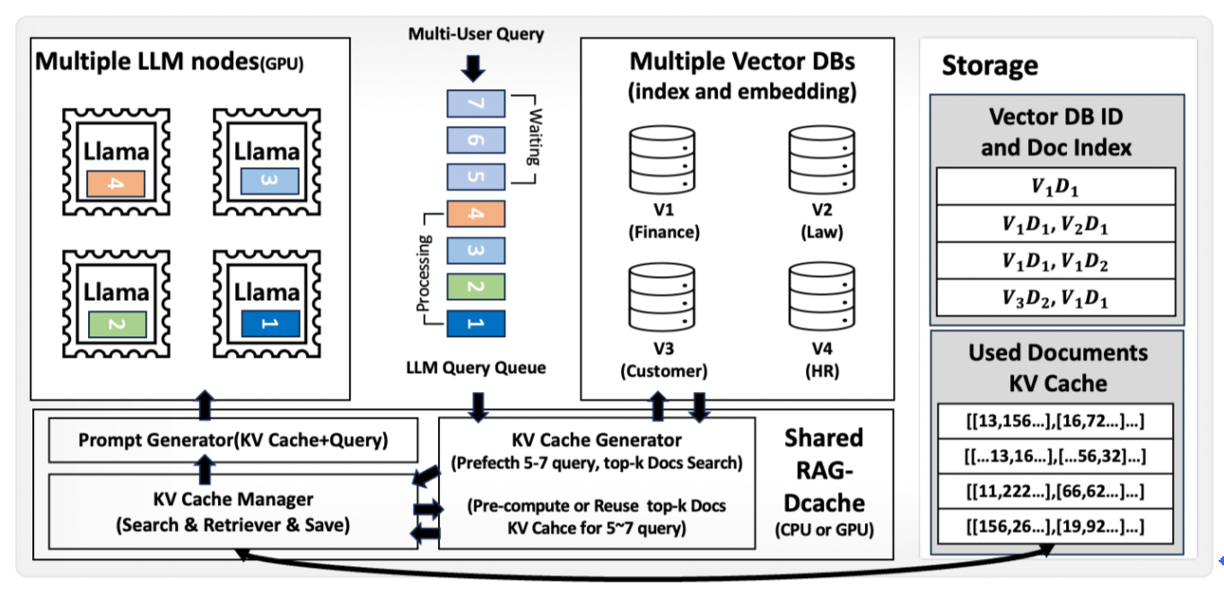}
    \caption{Shared RAG-DCache Architecture.}
    \label{fig:des_stsr}
\end{figure}

Figure~\ref{fig:des_stsr} depicts the architecture of Shared RAG-DCache, which builds on the single-instance design with additional components for multi-instance coordination. The main components are as follows:

\squishlist
\item 
\textbf{KV Cache Generator}: This is a new background module that proactively creates KV caches for queued queries. It continuously monitors the central query request queue and identifies queries that have been waiting longer than a predetermined threshold. For a query that is stuck in the queue (indicating the system is busy and the query will not be served immediately), the KV Cache Generator takes action: it immediately computes an embedding for the query, uses it to search the vector database for top-k relevant documents, and then computes the KV caches for those documents on the fly. Essentially, it performs steps 2–5 of the RAG-DCache workflow ahead of time for queries that are still waiting. The KV generation uses the same LLM model that will ultimately answer the query, but it can be executed on any available device – for example, on an idle GPU if one exists, or on the CPU if all GPUs are busy – since this is done asynchronously. Once generated, the new KV cache is stored to disk via the KV Cache Manager and indexed by document ID so that any LLM instance can retrieve it later. If a KV cache for a particular document was already created previously, the generator will detect this and avoid redundant computation. In that case, the existing cache can be reused directly. This component effectively prefetches document caches during the query’s waiting time, leveraging otherwise idle compute resources to reduce future work.

\item 
\textbf{Shared KV Cache Manager}: In a multi-instance setup, instead of each LLM instance having its own independent KV Cache Manager, we deploy a centralized KV Cache Manager service. This service coordinates the storage and sharing of KV caches among all LLM instances. It receives newly generated caches from the KV Cache Generator and inserts them into the global disk-based cache. When an LLM instance needs a KV cache for a document, it queries this shared manager rather than a local disk, and the manager supplies the data to the instance over the network or inter-process channel. The Shared KV Cache Manager thus acts as a cache server, ensuring that all LLM instances have a consistent view of available KV entries and that once a document’s cache is generated by any one instance or the generator, it can be used by all. Like the single-instance manager, it also implements a memory caching layer(using CPU RAM) with an eviction policy (e.g., LRU) to keep frequently accessed caches readily available. This is especially important in multi-instance scenarios to avoid repeatedly hitting the disk if multiple instances request the same cache around the same time. 

\item 
\textbf{Prompt Generator (per-instance)}: Each LLM instance is equipped with a Prompt Generator module. This component is conceptually similar to the RAG Processor’s prompt composition step in the single-instance case, but tailored for a multi-instance context. When a query is assigned to a specific LLM instance for processing, that instance’s Prompt Generator will request any needed KV caches from the Shared KV Cache Manager. It then combines the retrieved KV cache(s) with the query text to form the final input prompt for its local LLM, identical to how it was described in the single-instance workflow. 
With the KV cache preloaded into the model’s context, the LLM instance can skip directly to decoding the answer, greatly reducing the latency for that query. Essentially, the Prompt Generator ensures each instance makes full use of the globally cached data: it injects the shared KV into the model and thereby avoids that instance doing any redundant prefill computation for the documents.
\squishend

\begin{figure}[!t]
    \centering
    \includegraphics[width=1\linewidth]{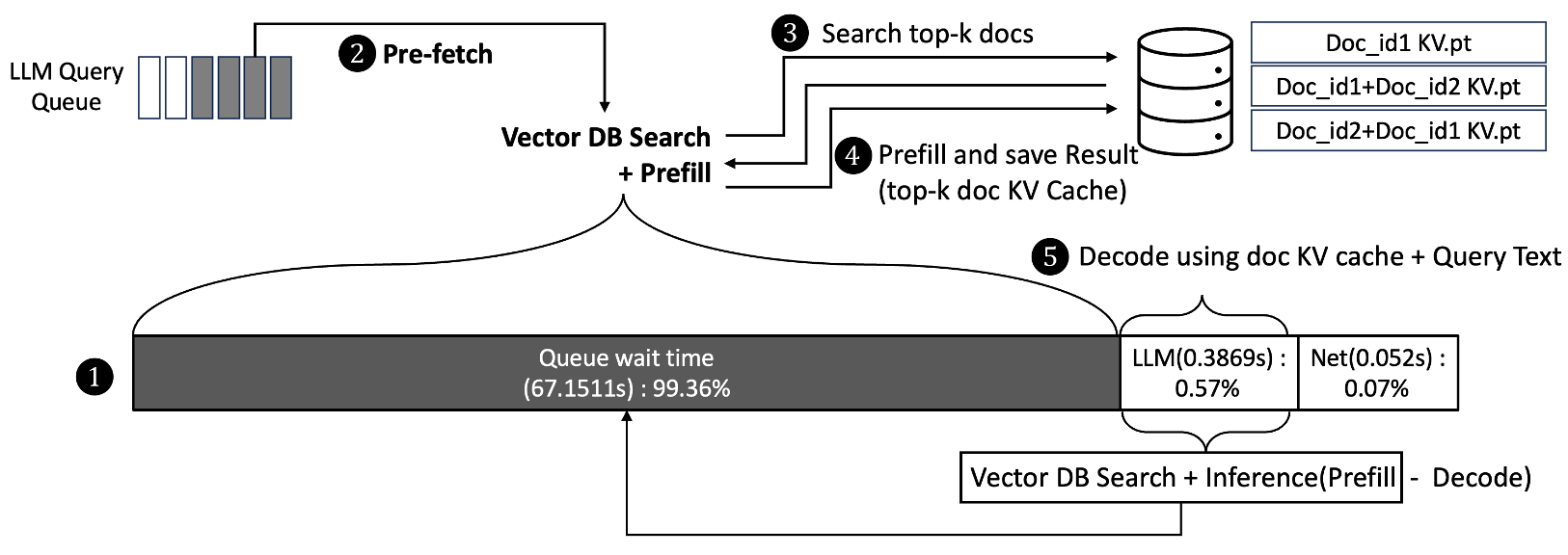}
    \caption{Shared RAG-DCache operation sequence.}
    \label{fig:des_sopt2}
\end{figure}

With these components, Shared RAG-DCache transforms a multi-instance deployment into a cooperative caching system. 
Figure~\ref{fig:des_sopt2} illustrates the Shared RAG-DCache operation sequence:

\bcircled{1} \bcircled{2}
Queue Monitoring: The system monitors the central queue of incoming queries continuously. If a query’s wait time exceeds a configured threshold (meaning the query has been in queue for a while due to heavy load), that query is flagged for cache prefetching. The threshold can be tuned – any query waiting longer is considered a good candidate to start processing early, since it likely will wait that long anyway.

\bcircled{3}
Document Pre-search: For each flagged query, the KV Cache Generator immediately kicks in. It takes the query, computes its embedding, and performs a vector DB lookup to fetch the top-k most similar documents. This step is analogous to the retrieval step normally done by an LLM instance, but here it happens in parallel, on an idle thread/CPU or a free GPU, while the query is still in queue. By doing this in advance, we obtain the set of documents we anticipate the query will need, without delaying the query’s actual service time.

\bcircled{4}
KV Cache Preparation: Next, for each of the k retrieved documents, the system prepares the KV cache. If the shared cache already contains a KV entry for a document, the KV Cache Generator will simply load that cache – possibly from disk to memory – immediately. If a cache is missing, the generator will perform a prefill computation for that document using the LLM model to create the KV cache. Once generated, the new KV cache is stored into the shared vector DB on disk via the Shared KV Cache Manager, making it available system-wide. This step effectively precomputes the heavy part of the LLM’s work for the document while the query is still waiting in line. It’s done for all top-k documents so that the query’s entire retrieved context is cached ahead of time.

\bcircled{5}
LLM Inference: Eventually, the queued query reaches the front of the queue and is assigned to an LLM instance for execution. At this point, thanks to the previous steps, the KV caches for its relevant documents have likely already been generated and stored. The assigned instance’s Prompt Generator fetches those caches from the Shared KV Cache Manager. The Prompt Generator then constructs the LLM prompt, combining the query text with the retrieved KV caches. Now the LLM can immediately begin decoding the answer, since the time-consuming document encoding work was done earlier. In effect, the query’s waiting time has been utilized to do useful computation, so when the query is actually processed, the response is much faster. A significant portion of the would-be latency has been shaved off, resulting in a substantially lower response time for the user.

Through this mechanism, Shared RAG-DCache ensures that no two instances ever duplicate the same KV computation, and that the query wait times in a busy service are put to productive use. Since all KV caches reside in a shared disk-based store, any cache generated by one instance or by the prefetcher is immediately available to all other instances. This not only cuts down latency for the individual query that triggered the prefetch, but also improves throughput overall: multiple LLM instances can pull from the same cache repository, benefitting from each other’s work. The design takes advantage of the locality in user queries and the nature of queued request handling to significantly reduce redundant computation across the system.

\section{Evaluation}

\begin{figure*}[ht!bp]  
\centering
    \includegraphics[width=0.80\textwidth]{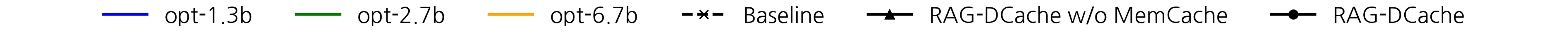} 
    \vspace{1pt} 
    \subfigure[TTFT]{
        \centering
        \includegraphics[width=0.26\textwidth]{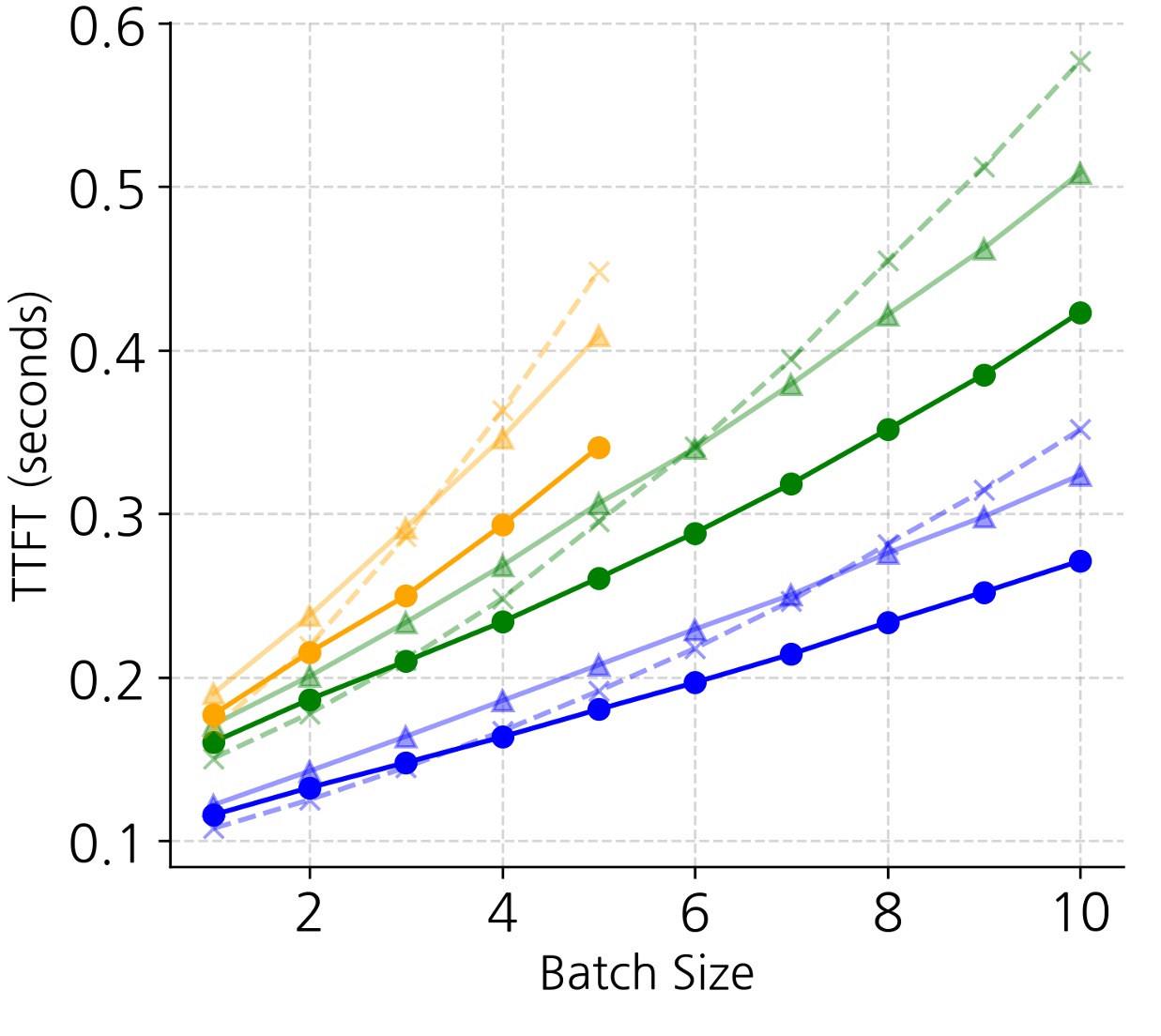}
        \label{fig:expr_a}
    }
    \hspace{0pt}
    \subfigure[Prefill Time]{
        \centering
        \includegraphics[width=0.26\textwidth]{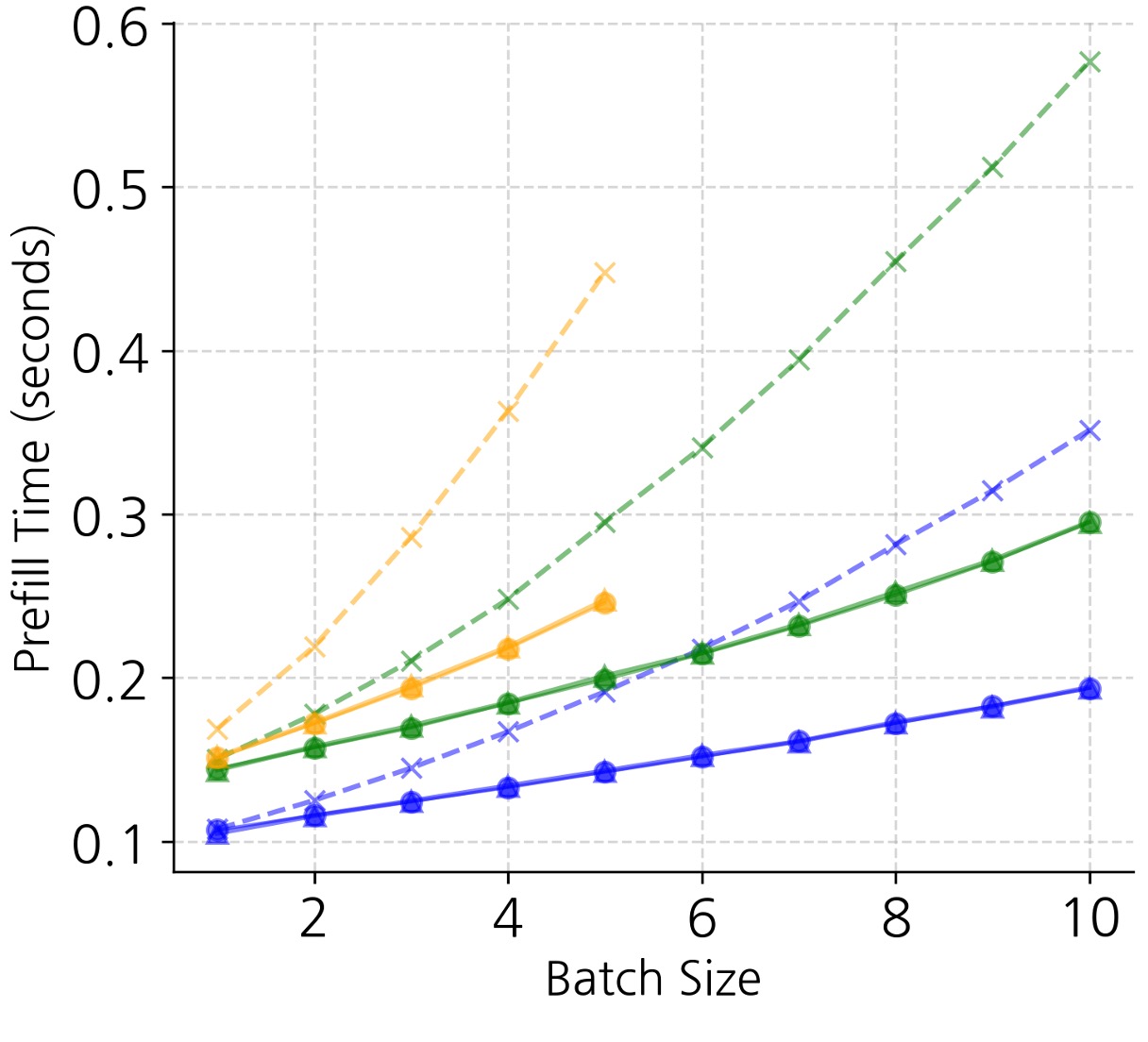}
        \label{fig:expr_b}
    }
    \hspace{0pt}
    \subfigure[KV Cache Loading Time]{
        \centering
        \includegraphics[width=0.26\textwidth]{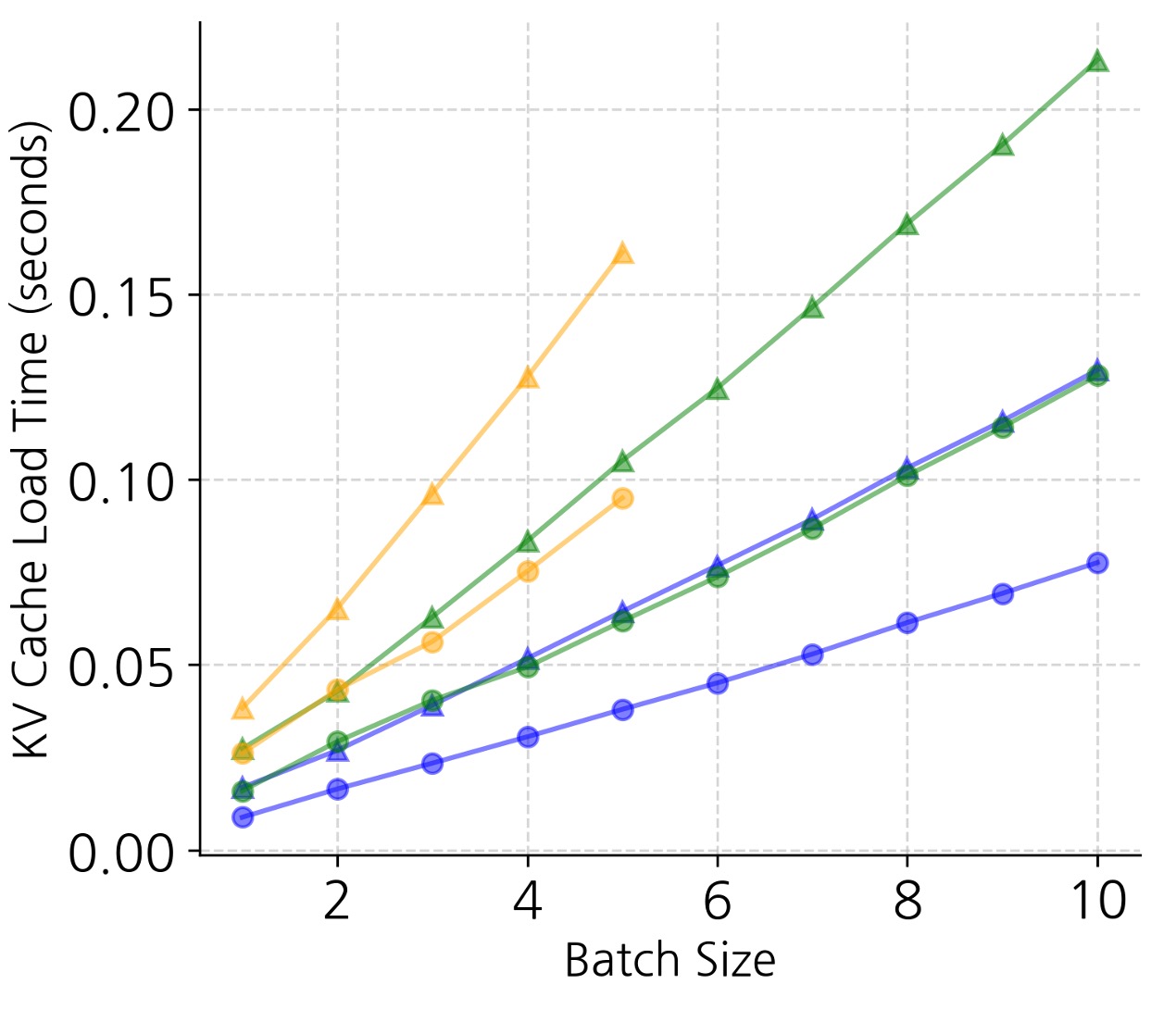}
        \label{fig:expr_c}
    }
    \caption{Results of TTFT, Prefill Time, and KV Cache Loading Time Measurements by LLM and Batch Size.
    }
    \label{fig:expr_rag}
\end{figure*}

We implemented the above system and conducted experiments to evaluate the performance gains of RAG-DCache and Shared RAG-DCache. All experiments were performed on a single-host server with the following specifications(Table~\ref{tab:tab0}):

\begin{table}[htbp]
    \centering
    \caption{Experimental hardware specifications.}
    \label{tab:tab0}
    \small
    \begin{tabular}{|l|p{6cm}|}
        \hline
        \textbf{Component} & \textbf{Specification} \\ \hline
        CPU & AMD Ryzen 9 3900XT (12 cores) \\ \hline
        GPU & NVIDIA RTX 2080 SUPER(2 units, 8GB each) \\ \hline
        Memory & 64GB RAM \\ \hline
        Storage & SAMSUNG 970 EVO NVMe SSD 500GB (read 3.4GB/s Write:2,4GB/s) \\ \hline
    \end{tabular}
\end{table}


We used the SQuAD v1.1 dataset as a source of queries and documents for retrieval, and a FAISS\cite{johnson2021billion} vector database for similarity search. Different large language models were used for single-instance vs. multi-instance tests, and we measured key metrics including TTFT and throughput (queries processed per second), as well as breakdowns of latency where appropriate. TTFT captures the latency from when a query is submitted to when the LLM outputs the first token of the answer – this primarily reflects the time spent in the prefill stage since decoding the very first token is usually quick once the model has the prompt. Throughput is measured as the number of queries that can be completed per second, reflecting the system’s capacity under load.

The increase in the vector database storage capacity due to disk-based KV cache usage is shown in Table~\ref{tab:tab4}. The FAISS Vector DB size is the sum of the index, the document embedding vectors, and the original text size of the documents. RAG-DCache and Shared RAG-DCache require additional disk space for storing KV caches in addition to the FAISS vector database storage. As expected, the KV cache size increases with the number of model parameters. Even with the same number of parameters, the KV Cache size can differ depending on the embedding method used by each LLM model.
Additionally, if the number of documents extracted in RAG (top-k) increases, the size of the KV Cache generated may also grow. However, by adjusting the disk-based KV Cache that leverages query-document locality, the required storage space can be reduced—this may represent a tradeoff between query throughput and the necessary disk size.


\begin{table}[htbp]
\centering
\caption{Size of Normal FAISS Vector DB and KV Cache when using SQuAD dataset and different LLMs.}
\fontsize{9.5pt}{11pt}\selectfont 
\begin{tabular}{|c|c|c|c|c|} 
\hline  
  \multirow{2}{*}{\makecell{\textbf{FAISS} \\ \textbf{Vector DB}}} & \multicolumn{4}{c|}{\textbf{RAG-DCache and Shared RAG-DCache}} \\ \cline{2-5}
   & \textbf{opt-1.3b} & \textbf{opt-2.7b} & \textbf{opt-6.7b} & \textbf{LLAMA-1B} \\ \hline 
  0.5MB & 5.9GB & 9.9GB & 16GB & 1GB \\ \hline 
\end{tabular}
\label{tab:tab4}
\end{table}


\subsection{RAG-Dcache Results and Analysis}
For the single-instance scenario, we evaluated RAG-DCache using the SQuAD dataset and Facebook’s OPT\cite{zhang2022opt} decoder-only LLM of varying sizes. We tested three model sizes to see how cache benefits scale with model complexity. The vector database was implemented with Faiss, and we chunked the SQuAD documents into passages for retrieval. Before inference, we pre-generated the KV caches for all document chunks that might be retrieved for the SQuAD queries, by running each chunk through the respective OPT model’s prefill stage and storing the resulting KV pairs on disk. The KV Cache Manager was given a 16GB memory cache to hold recently used caches, as described earlier. Table~\ref{tab:tab1} summarizes the experimental setup, including the LLM models, dataset size (2,000 queries from SQuAD v1.1 train), and other components. We compared two settings: a baseline RAG (meaning the LLM processes raw text of retrieved documents for every query) and RAG-DCache enabled, across different batch sizes. Here, “batch size” refers to the number of queries processed simultaneously by the model. For each combination of model size and batch size, we measured the average TTFT and the throughput in both the baseline and RAG-DCache configuration.

\begin{table}[htbp]
    \centering
    \caption{Experimental Environment.}
    \label{tab:tab1}
    \small
    \begin{tabular}{|l|l|}
        \hline
        \textbf{Component} & \textbf{Specification} \\ \hline
        LLM & facebook/opt-1.3b, 2.7b, 6.7b \\ \hline
        Embedding Model & all-MiniLM-L6-v2 \\ \hline
        Dataset & SQuAD v1.1 Train (2,000 samples) \\ \hline
        Vector DB & Faiss DB, IndexFlatIP \\ \hline
    \end{tabular}
\end{table}

\textbf{Performance Breakdown}: Figure~\ref{fig:expr_rag} presents the results for TTFT, Prefill time, and KV cache loading time under various conditions. Figure~\ref{fig:expr_a} shows the overall TTFT for each model and batch size, comparing the baseline to RAG-DCache. Figure~\ref{fig:expr_b} and Figure~\ref{fig:expr_c} break this TTFT into two components for the RAG-DCache case: the time spent in the LLM’s prefill stage, and the time spent loading KV caches from disk. In the baseline, TTFT is essentially all prefill. 

As shown in Figure~\ref{fig:expr_a}, RAG-DCache consistently reduces TTFT in almost all cases. This is because using the disk-based KV cache drastically reduces the prefill computation time on the GPU, and the memory cache further cuts down repeated disk reads. In Figure~\ref{fig:expr_b}, we see that the prefill time with RAG-DCache is much lower than baseline since the LLM doesn’t need to encode the full documents from scratch. Figure~\ref{fig:expr_c} shows the KV cache loading time incurred for RAG-DCache – this is an overhead not present in the baseline. However, because of our caching optimizations, this overhead is kept relatively small: many cache loads are served from the 16GB memory cache, and even disk loads are fast on NVMe SSD. The result is that Prefill time savings outweigh the KV load time, yielding a net gain. For example, with the OPT-6.7B model at batch size 4, RAG-DCache might add a few milliseconds to load caches but saves far more time in GPU computation, leading to a substantially lower TTFT overall.

\begin{figure}[htbp]
\centerline{\includegraphics[width=\linewidth]{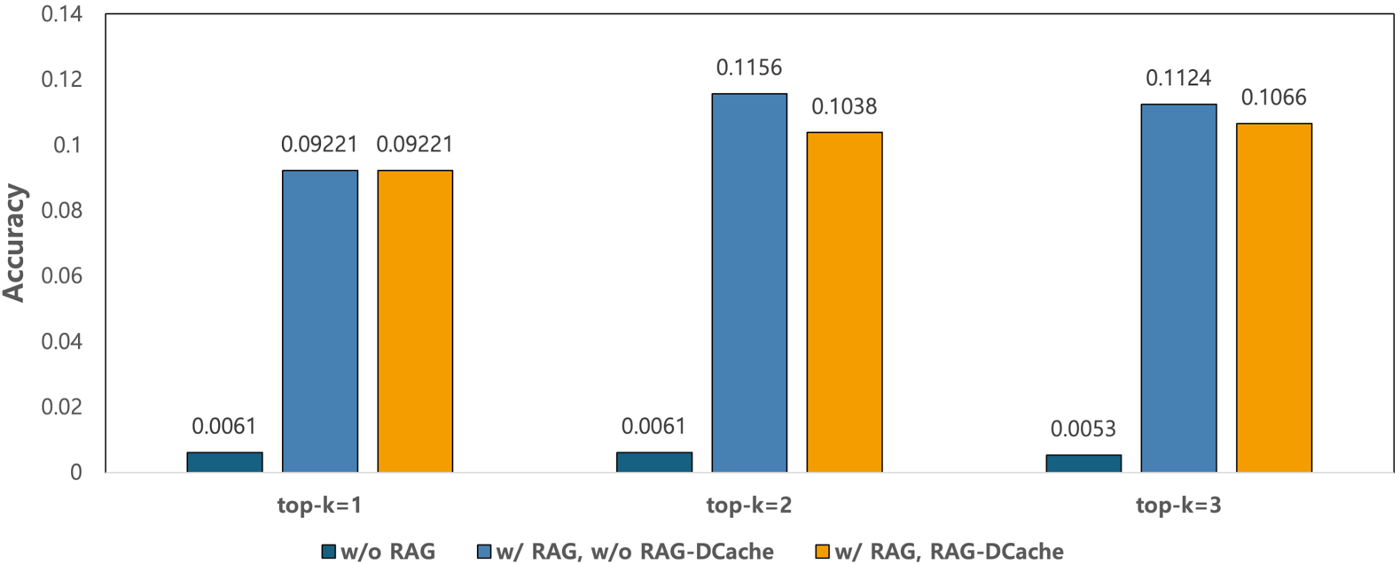}}
\caption{Accuracy based on the number of retrieved documents (top-k) when using RAG-DCache.}
\label{fig:exp_accu}
\end{figure}

However, as shown in the Figure~\ref{fig:exp_accu} RAG-DCache works well when only a single document(top-k=1) is retrieved. We measured the accuracy of the answer using the formula F1 × 0.5 + Exact Match × 0.5, and the results showed that when top-k=1, the accuracy was the same as when RAG-DCache was not used. However, when multiple documents (top-k$>$1) were retrieved, the accuracy was higher than when RAG was not used but lower than when RAG-DCache was not applied. This is because we do not calculate cross-attention between the documents.

\textbf{Impact of Model Size}: The benefits of RAG-DCache become more pronounced for larger models and larger batch sizes. Larger models have heavier per-token computation, so removing a chunk of tokens from their workload yields a bigger absolute time savings. Similarly, at higher batch sizes, the GPU is encoding multiple queries’ documents at once in the baseline, which is very compute-intensive; with caching, those computations are skipped, freeing the GPU to handle more queries. 

In our experiments, we observed TTFT reductions of roughly 10\%–20\% with RAG-DCache compared to the baseline, depending on the configuration. These percentages tended towards the higher end (closer to 20\%) for larger OPT models and larger batches. Concretely, Table~\ref{tab:tab2} shows the throughput achieved in each scenario, averaged per model. 
Without RAG-DCache, the throughput for OPT-1.3B was approximately 23.74 QPS, which increased to 26.63 QPS with RAG-DCache – approximately a 12\% improvement. 
For the OPT-2.7B model, throughput went from 15.32 QPS to 18.01 QPS, approximately a 17.6\% gain. The OPT-6.7B model was the slowest, but improved to 11.05 QPS with caching (15.7\% increase). The average relative improvement in these models was around 14\%–15\%, aligning well with the TTFT savings noted above. These results validate that RAG-DCache not only lowers the latency per query but also increases the overall throughput of the system, as the GPUs spend less time on redundant tasks and can handle more queries. 

\begin{table}[htbp]
\centering
\caption{Average Throughput of Baseline and RAG-DCache.}
\fontsize{9.5pt}{11pt}\selectfont 
\begin{tabular}{|c|c|c|c|c|} \hline  
  \textbf{Component}& \textbf{opt-1.3b} & \textbf{opt-2.7b} & \textbf{opt-6.7b} &  \textbf{Average}\\ \hline 

 \textbf{Baseline} & 23.74 & 15.32 & 9.55 & 17.53 \\ \hline 
 \textbf{RAG-DCache} & 26.63 & 18.01 & 11.05 & 20.07 \\ \hline
\end{tabular}
\label{tab:tab2}
\end{table}


\subsection{Shared RAG-DCache Results and Analysis}
We next evaluate Shared RAG-DCache in a multi-instance LLM service scenario. The test environment remained the same dual-GPU server described above. We switched the LLM to Meta’s LLaMA-3.2-1B model for these experiments. 
We used 1,000 queries from the SQuAD v1.1 dataset and for each query.
We retrieved either k=1 or k=2 documents to examine the effect of different context sizes. 
These queries were sent in a continuous stream at a rate of 40 requests per second to simulate a heavy multi-user load. This high query rate ensured that at any given moment there were multiple queries waiting in the queue, which is necessary to fully leverage the prefetching mechanism of Shared RAG-DCache. If the system is not under load, queries won’t wait in queue and the KV Cache Generator might not trigger. 

And whether Shared RAG-DCache was used or not, the KV cache values generated during the prefill stage remained the same regardless of changes in the top-k value. 
Therefore, accuracy was not measured separately. Furthermore, to control for variability in the decode phase, this experiment focused solely on measuring the prefill stage. This is because response latency can vary depending on the length of the generated answer during the decode stage. Our goal was to reduce noise and highlight the optimization benefits of the prefill stage. 
And to isolate the effect of shared disk caching, we disabled the memory caching in the KV Cache Manager for these experiments. 
This means all cache fetches go to disk, ensuring that any performance improvements observed are due to the multi-instance sharing and prefetching, not just RAM hits. Finally, to measure the performance improvement effect as the Disk-based KV Cache exapnds, 
we processed the entire dataset multiple times in a random order.

\begin{figure}[!t]
\centerline{\includegraphics[width=\linewidth]{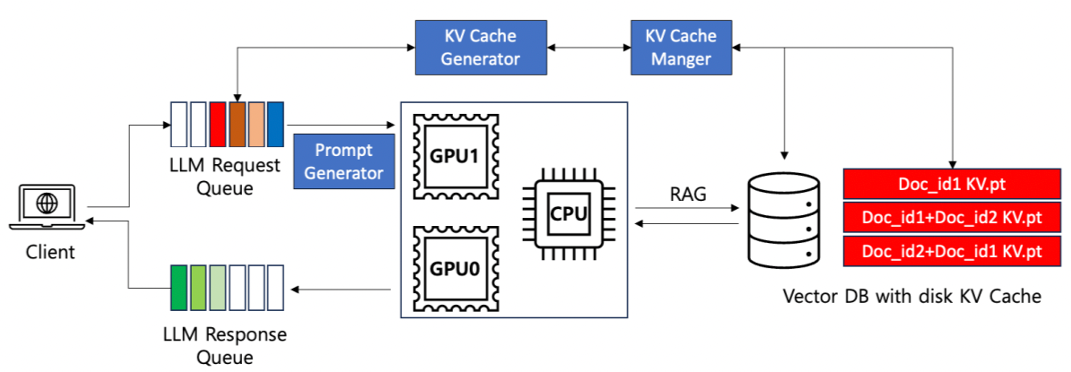}}
\caption{Experimental System Configuration.}
\label{fig:exp_conf}
\end{figure}

The test configuration for this experiment is shown in Figure~\ref{fig:exp_conf}. To determine the optimal configuration in the given environment, we evaluated two resource allocation strategies for CPU and GPU. 
\squishlist
\item
In Configuration (A) “GPU-Only KV Cache Generation”, we dedicate one of the two GPUs entirely to KV cache generation tasks, and use only the other GPU to run the LLM inference for answering queries. In other words, GPU0 handles all LLM inference requests, and GPU1 is reserved for computing KV caches of retrieved documents in the background. 
\item
In Configuration (B) “CPU-Based KV Cache Generation”, we use both GPUs for LLM inference, and assign all KV cache generation to the CPU. In this setup, the KV Cache Generator runs on the CPU, while both GPU0 and GPU1 are busy serving LLM inference requests. 
\squishend

Configuration (A) tests the scenario where we sacrifice a GPU to speed up cache prep, whereas (B) tests using no GPU for cache prep at the expense of slower cache generation on CPU. In both cases, Shared RAG-DCache is active, which means that caches are shared across the two LLM instances, and prefetching is enabled. The baseline for comparison is a multi-instance system without Shared RAG-DCache(w/o KVGen). We measure the system’s throughput in queries/sec and the average end-to-end latency which in our prefill-only measurement corresponds to how long a query waits plus its prefill time for each configuration and each top-k documents.

\begin{figure}[!t]
    \centering
    \subfigure[top-k=1]{
        \includegraphics[width=1\linewidth]{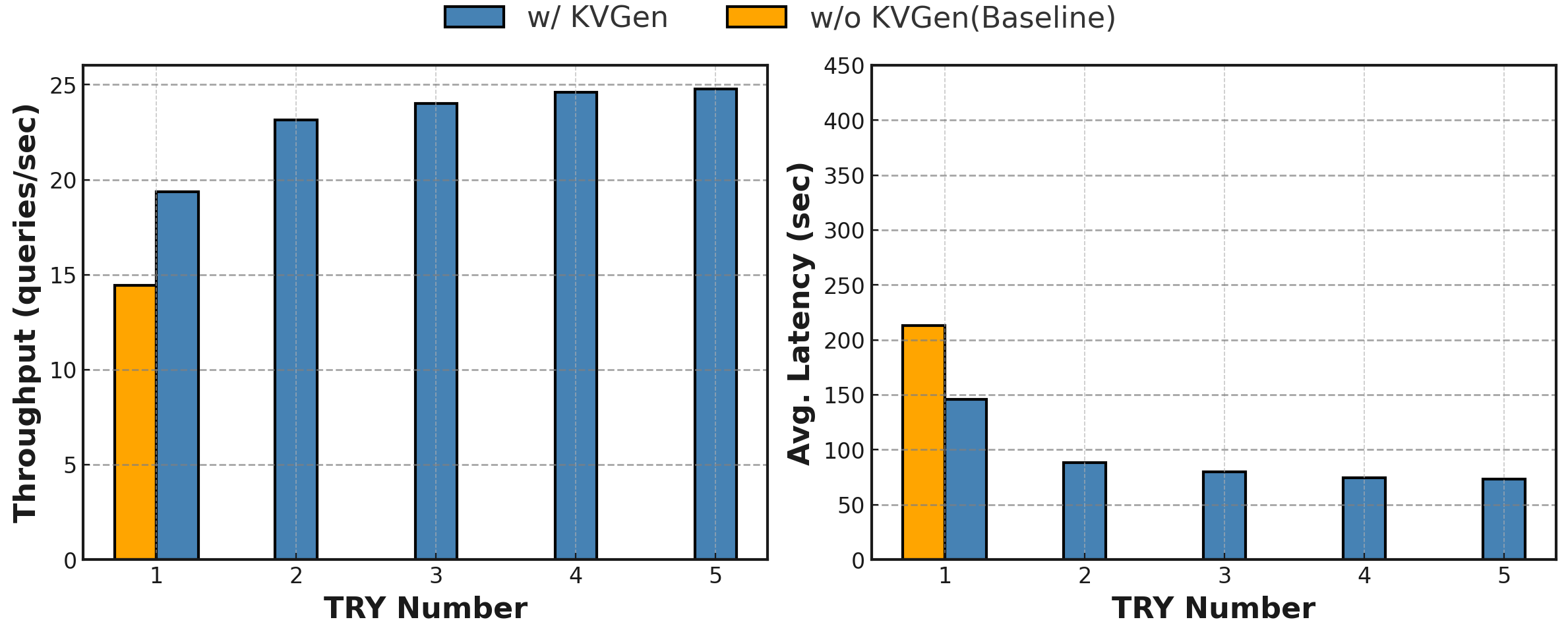}
        \label{fig:expr_f9-1}
    }
    \hfill
    \subfigure[top-k=2]{
        \includegraphics[width=1\linewidth]{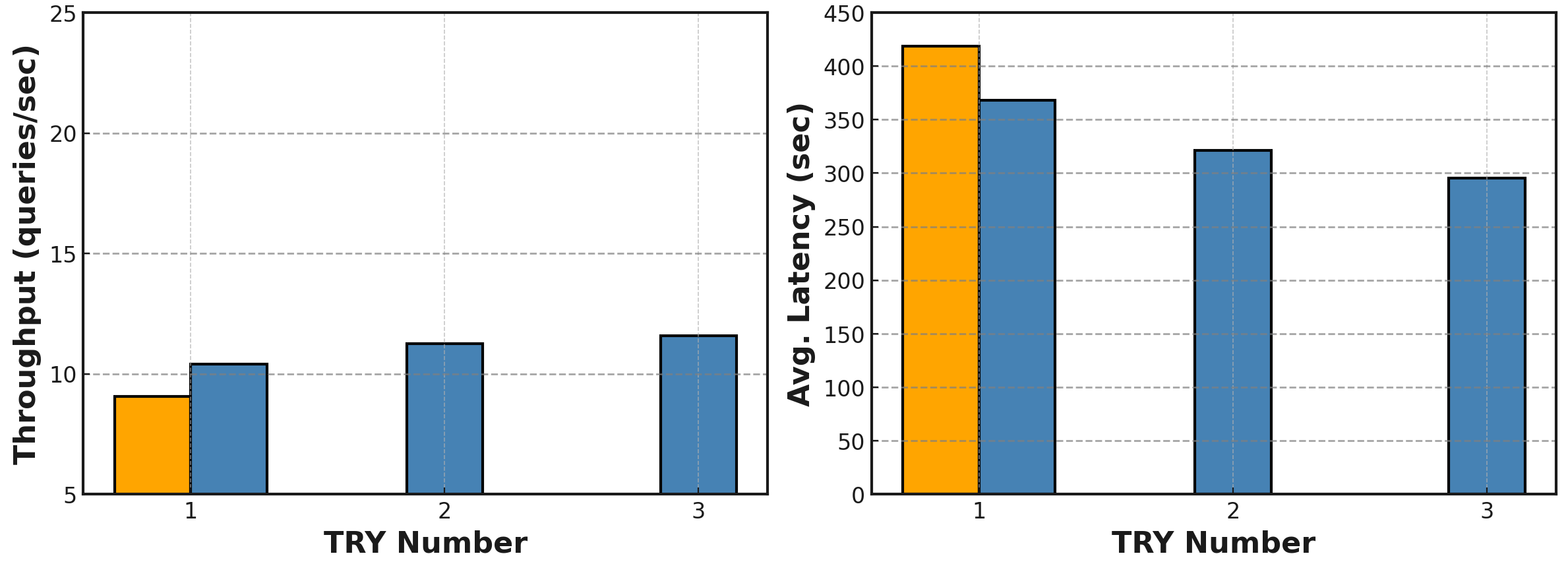}
        \label{fig:expr_f9-2}
    }
    \caption{Throughput and Avg. Latency with Configuration (A) when top-k=1 and 2.}
    \label{fig:exp_fig9}
\end{figure}

\textbf{Overall Improvements}: Shared RAG-DCache showed significant performance improvements over the baseline, though the magnitude depended on the configuration. In Configuration (A), as shown in Figure ~\ref{fig:exp_fig9}, enabling the shared cache system to improve throughput by approximately 35\%-71\% and reduce average latency by 31\%–65\% compared to the baseline, according to our measurements.
As the number of times the dataset was processed(TRY Number in the figure) increased, the performance improved further.(For the baseline, since there is no disk-based KV Cache expansion according to the number of tries (as with Shared RAG-DCache), it was measured only once.)

For a example, in the k=1 scenario under Configuration (A), throughput increased from 23.96 QPS to 24.78 QPS with caching, and the average latency dropped from 75.75s to 73.25s. These particular numbers represent a modest 3.4\% throughput gain and 3.3\% latency reduction – relatively small, because with only one document the baseline was already not very slow and the single inference GPU was not heavily bottlenecked.

In the k=2 scenario with Configuration (A), we observed mixed results: the caching system sometimes incurred overhead that offset its benefit. Specifically, handling two documents per query on only one inference GPU proved challenging – the throughput and latency with caching in some trials were on par with or slightly worse than the baseline. This is likely because in Configuration (A) the single GPU had to handle the combined work of two documents’ KV insertion plus the query itself, sequentially, which increased contention. The KV generation GPU could produce caches quickly, but the inference GPU became a bottleneck when k was larger. 
However, as the TRY Number, which represents the number of times the dataset was repeatedly processed, increased, throughput improved, and latency decreased progressively. This is attributed to the accumulation of the Disk KV Cache, which increased the likelihood that queries processed by the LLM would reference the same KV Cache, thereby improving the Disk KV Cache hit ratio.

In summary, Configuration (A) demonstrated the viability of shared caching, but it showed limited parallelism – dedicating a GPU for caches helped less than expected when the remaining GPU was overloaded with inference work for larger contexts. However, as the Disk KV Cache accumulated, throughput improved, and latency progressively decreased.

\begin{figure}[!t]
    \centering
    \subfigure[top-k=1]{
        \includegraphics[width=1\linewidth]{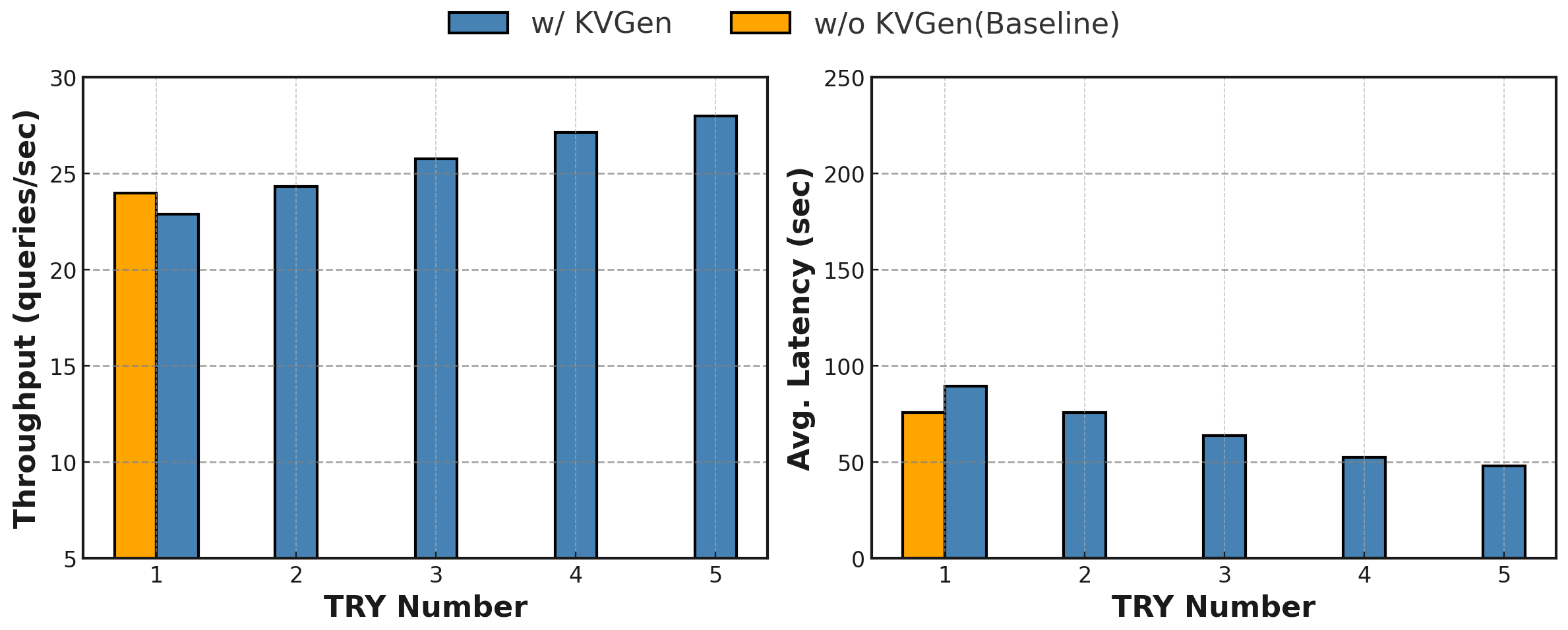}
        \label{fig:expr_f10-1}
    }
    \hfill
    \subfigure[top-k=2]{
        \includegraphics[width=1\linewidth]{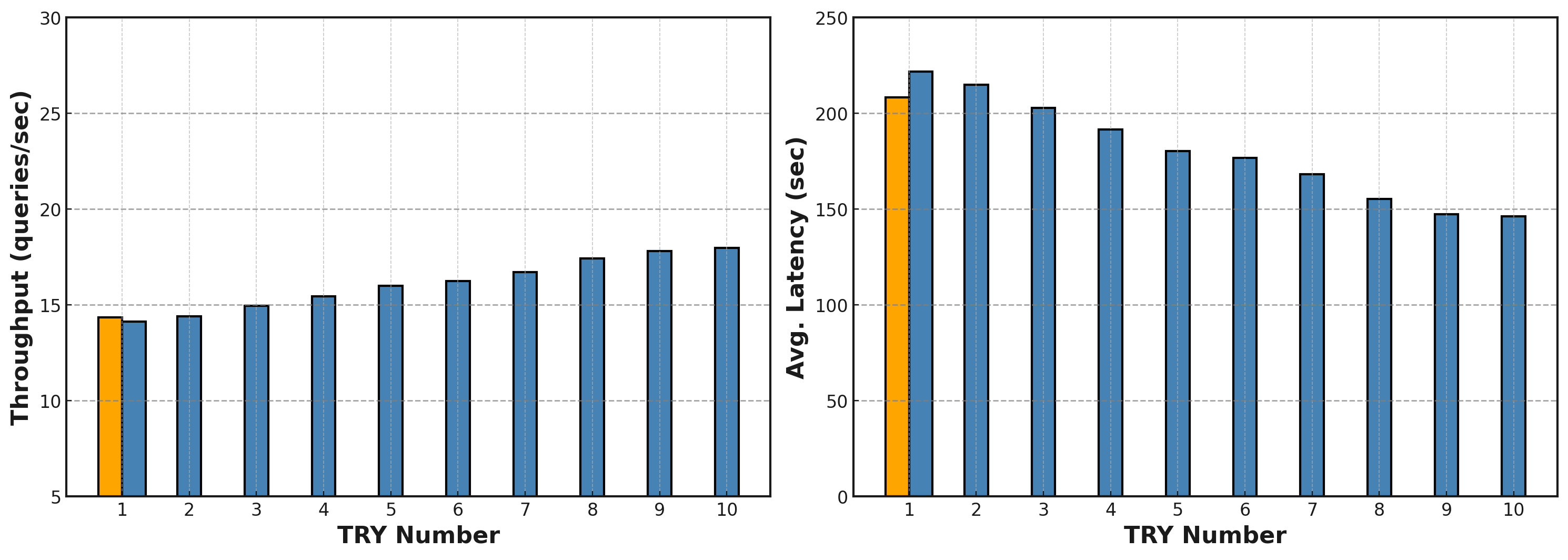}
        \label{fig:expr_f10-2}
    }
    \caption{Throughput and Avg. Latency with Configuration (B) when top-k=1 and 2. }
    \label{fig:exp_fig10}
\end{figure}

In Configuration (B), as shown in Figure~\ref{fig:exp_fig10}, with caching enabled, overall throughput increased by approximately 15\%-28\% and latency decreased by 12\%–29\% compared to baseline in this configuration. For example, as shown in Fig.10, in the k=1 case, as the number of times the dataset was processed increased, the throughput improved from 23.96 to 27.98 QPS, and the average latency fell from 75.75s to 47.92s. This is a significant improvement: ~17\% higher throughput and ~37\% lower latency. For k=2, the system with caching went from 14.34 QPS to 17.96 QPS, and latency dropped from 208.22s to 146.17s. That’s roughly a 25\% increase in throughput and a 30\% reduction in latency, respectively.

We note that k=2 queries are inherently slower – even with caching, the latency was higher than any k=1 scenario simply because processing two documents’ worth of context takes extra time and resources. However, the relative improvement with caching is still substantial for k=2. These results confirm that Shared RAG-DCache is effective even when additional context is included, although the best absolute performance naturally occurs with fewer documents. Indeed, comparing k=1 vs k=2 across the board, we see that k=1 had lower latency and higher throughput in all configurations (baseline and caching). This is expected because more documents means more work. 

Importantly, however, Shared RAG-DCache mitigates the performance penalty of larger k: for instance, going from 1 to 2 documents in the baseline caused throughput to drop by ~40\% and latency to almost triple, whereas with Shared RAG-DCache the drop in throughput was less severe and the latency increase was much smaller – and some of that remaining latency was due to the heavier workload rather than idle waiting. Even with k=2, the caching system delivered significantly better performance than baseline.

\textbf{Comparison of Configurations (A) vs (B)}: The comparison result is shown in Table~\ref{tab:tab3}. The CPU-based KV generation (B) clearly emerges as the preferable configuration in our experiments. It achieved the highest throughput in all cases and more consistent latency reductions. In Configuration (A), when one GPU was taken away from inference, the remaining single inference GPU became a choke point under heavier loads (like k=2). It could not parallelize the work enough, and the benefit of offloading some computation to the second GPU was negated by the loss of overall inference capacity. In configuration (B), both GPUs were fully utilized for inference tasks, doubling the inference parallelism, while the CPU handled cache generation without impeding the GPUs. The CPU was effectively leveraging otherwise idle time since GPUs were busy, CPU cycles could be used to precompute caches. This leads to better pipeline balance: GPU power focused on what GPUs do best running the model for answers, and CPU cycles used for background prep work. 

As a result, configuration (B) achieved the highest observed throughput in our tests – for example, ~27.98 QPS at k=1 with caching, which was higher than even the baseline with two GPUs. It even outperformed Configuration (A)’s throughput despite Configuration (A) having a whole GPU doing cache work, indicating that that GPU might have been under-utilized or its benefit was offset by the other GPU’s overload. Additionally, the latency in configuration (B) was dramatically better: in Table III, configuration (B) brought average latency down to ~47.9s for k=1 and 146.2s for k=2, whereas Configuration (A) had 73.3s (k=1) and a very high 295.4s . 

The k=2 result for Configuration (A) (295s) suggests that the single GPU was so overwhelmed that queries ended up waiting a long time despite caching – possibly because the KV generation GPU was producing caches faster than the inference GPU could use them, leading to a queue buildup. In contrast, configuration (B) kept latencies much lower by always utilizing both GPUs for serving queries. These findings underscore that offloading KV generation to a non-GPU resource yields better overall system performance, which aligns with our resource allocation optimization strategy. 

\begin{table}[h]
\centering
\caption{Average Performance Results by Configuration}
\resizebox{\linewidth}{!}{%
\begin{tabular}{|c|c|c|c|c|}
\hline
Top-k & Avg. Perf. & Conf. (A) & GPU: 2 LLM(Baseline) & Conf. (B) \\ 
\hline
\multirow{2}{*}{1} & Throughput & 24.78 & 23.96 & \textbf{27.98} \\ 
\cline{2-5} 
                   & Latency    & 73.25 & 75.75 & \textbf{47.92} \\ 
\hline
\multirow{2}{*}{2} & Throughput & 11.57 & 14.34 & \textbf{17.96} \\ 
\cline{2-5} 
                   & Latency    & 295.37 & 208.22 & \textbf{146.17} \\ 
\hline
\end{tabular}
}
\label{tab:tab3}
\end{table}

In summary, Shared RAG-DCache provided some performance improvements in both configurations compared to the baseline, Utilizing not only GPU but also CPU resources for cache generation can maximize effectiveness. Our optimal setup in this experiment was configuration (B), which improved throughput by ~16.8\% and ~25.2\% for k=1 and 2 respectively, and cut latencies by ~36.8\% and ~29.8\%, relative to the no-caching baseline as summarized in Table~\ref{tab:tab3}. Configuration (A) was suboptimal, showing the importance of a balanced resource allocation when using Shared RAG-Dcache.


\section{Conclusion}
In this paper, we proposed and implemented a disk-based shared KV cache management system—\textit{Shared RAG-DCache}—to optimize LLM inference in multi-instance service environments. 
By leveraging query locality and service queue waiting times, our approach prefetches and shares the KV caches of frequently accessed documents across multiple instances. This significantly reduces redundant prefill computations and increases overall throughput while reducing response latency. 
Experiments on a dual-GPU/one-CPU server demonstrated throughput improvements of up to 70\% and latency reductions of tens of percentage points. 
In particular, an optimal configuration was achieved when all GPUs were dedicated to LLM inference and the CPU handled KV cache generation, ensuring maximal performance gains.


\footnotesize{ 
    \bibliographystyle{ieeetr}
    \bibliography{ref} 

\begin{thebibliography}{10}

\bibitem{10.5555/3495724.3496517}
P.~Lewis, E.~Perez, A.~Piktus, F.~Petroni, V.~Karpukhin, N.~Goyal, H.~K\"{u}ttler, M.~Lewis, W.-t. Yih, T.~Rockt\"{a}schel, S.~Riedel, and D.~Kiela, ``Retrieval-augmented generation for knowledge-intensive nlp tasks,'' in {\em Proceedings of the 34th International Conference on Neural Information Processing Systems}, NIPS '20, (Red Hook, NY, USA), Curran Associates Inc., 2020.

\bibitem{siriwardhana-etal-2023-improving}
S.~Siriwardhana, R.~Weerasekera, E.~Wen, T.~Kaluarachchi, R.~Rana, and S.~Nanayakkara, ``Improving the domain adaptation of retrieval augmented generation ({RAG}) models for open domain question answering,'' {\em Transactions of the Association for Computational Linguistics}, vol.~11, pp.~1--17, 2023.

\bibitem{chen2023benchmarkinglargelanguagemodels}
J.~Chen, H.~Lin, X.~Han, and L.~Sun, ``Benchmarking large language models in retrieval-augmented generation,'' 2023.

\bibitem{fu2024lazyllm}
Q.~Fu, M.~Cho, T.~Merth, S.~Mehta, M.~Rastegari, and M.~Najibi, ``Lazy{LLM}: Dynamic token pruning for efficient long context {LLM} inference,'' in {\em Workshop on Efficient Systems for Foundation Models II @ ICML2024}, 2024.

\bibitem{vaswani2017attention}
A.~Vaswani, ``Attention is all you need,'' {\em Advances in Neural Information Processing Systems}, 2017.

\bibitem{NEURIPS2022_67d57c32}
T.~Dao, D.~Fu, S.~Ermon, A.~Rudra, and C.~R\'{e}, ``Flashattention: Fast and memory-efficient exact attention with io-awareness,'' in {\em Advances in Neural Information Processing Systems} (S.~Koyejo, S.~Mohamed, A.~Agarwal, D.~Belgrave, K.~Cho, and A.~Oh, eds.), vol.~35, pp.~16344--16359, Curran Associates, Inc., 2022.

\bibitem{Jin2024RAGCacheEK}
C.~Jin, Z.~Zhang, X.~Jiang, F.~Liu, X.~Liu, X.~Liu, and X.~Jin, ``Ragcache: Efficient knowledge caching for retrieval-augmented generation,'' {\em ArXiv}, vol.~abs/2404.12457, 2024.

\bibitem{Lu2024TurboRAGAR}
S.~Lu, H.~Wang, Y.~Rong, Z.~Chen, and Y.~Tang, ``Turborag: Accelerating retrieval-augmented generation with precomputed kv caches for chunked text,'' {\em ArXiv}, vol.~abs/2410.07590, 2024.

\bibitem{rajpurkar2016squad}
P.~Rajpurkar, J.~Zhang, K.~Lopyrev, and P.~Liang, ``Squad: 100,000+ questions for machine comprehension of text,'' in {\em Proceedings of the 2016 Conference on Empirical Methods in Natural Language Processing}, pp.~2383--2392, Association for Computational Linguistics, 2016.

\bibitem{10.1145/3600006.3613165}
W.~Kwon, Z.~Li, S.~Zhuang, Y.~Sheng, L.~Zheng, C.~H. Yu, J.~Gonzalez, H.~Zhang, and I.~Stoica, ``Efficient memory management for large language model serving with pagedattention,'' in {\em Proceedings of the 29th Symposium on Operating Systems Principles}, SOSP '23, (New York, NY, USA), p.~611–626, Association for Computing Machinery, 2023.

\bibitem{Wu2023FastDI}
B.~Wu, Y.~Zhong, Z.~Zhang, G.~Huang, X.~Liu, and X.~Jin, ``Fast distributed inference serving for large language models,'' {\em ArXiv}, vol.~abs/2305.05920, 2023.

\bibitem{xiao2024efficient}
G.~Xiao, Y.~Tian, B.~Chen, S.~Han, and M.~Lewis, ``Efficient streaming language models with attention sinks,'' in {\em The Twelfth International Conference on Learning Representations}, 2024.

\bibitem{280922}
G.-I. Yu, J.~S. Jeong, G.-W. Kim, S.~Kim, and B.-G. Chun, ``Orca: A distributed serving system for {Transformer-Based} generative models,'' in {\em 16th USENIX Symposium on Operating Systems Design and Implementation (OSDI 22)}, (Carlsbad, CA), pp.~521--538, USENIX Association, July 2022.

\bibitem{298687}
Y.~Zhong, S.~Liu, J.~Chen, J.~Hu, Y.~Zhu, X.~Liu, X.~Jin, and H.~Zhang, ``{DistServe}: Disaggregating prefill and decoding for goodput-optimized large language model serving,'' in {\em 18th USENIX Symposium on Operating Systems Design and Implementation (OSDI 24)}, (Santa Clara, CA), pp.~193--210, USENIX Association, July 2024.

\bibitem{rasley2022deepspeed}
J.~Rasley, O.~Ruwase, S.~He, S.~Ye, G.~Huang, J.~Liu, and Y.~Wang, ``Deepspeed inference: Enabling efficient inference of transformer models at unprecedented scale,'' {\em arXiv preprint arXiv:2207.00032}, 2022.

\bibitem{karpukhin-etal-2020-dense}
V.~Karpukhin, B.~Oguz, S.~Min, P.~Lewis, L.~Wu, S.~Edunov, D.~Chen, and W.-t. Yih, ``Dense passage retrieval for open-domain question answering,'' in {\em Proceedings of the 2020 Conference on Empirical Methods in Natural Language Processing (EMNLP)} (B.~Webber, T.~Cohn, Y.~He, and Y.~Liu, eds.), (Online), pp.~6769--6781, Association for Computational Linguistics, Nov. 2020.

\bibitem{8733051}
J.~Johnson, M.~Douze, and H.~Jégou, ``Billion-scale similarity search with gpus,'' {\em IEEE Transactions on Big Data}, vol.~7, no.~3, pp.~535--547, 2021.

\bibitem{yang2018hotpotqa}
Z.~Yang, P.~Qi, S.~Zhang, Y.~Bengio, W.~W. Cohen, R.~Salakhutdinov, and C.~D. Manning, ``Hotpotqa: A dataset for diverse, explainable multi-hop question answering,'' in {\em Proceedings of the 2018 Conference on Empirical Methods in Natural Language Processing}, pp.~2369--2380, 2018.

\bibitem{joshi2017triviaqa}
M.~Joshi, E.~Choi, D.~S. Weld, and L.~Zettlemoyer, ``Triviaqa: A large scale distantly supervised challenge dataset for reading comprehension,'' in {\em Proceedings of the 55th Annual Meeting of the Association for Computational Linguistics (Volume 1: Long Papers)}, pp.~1601--1611, 2017.

\bibitem{wang2020minilmv2}
W.~Wang, H.~Bao, S.~Huang, L.~Dong, and F.~Wei, ``Minilmv2: Multi-head self-attention relation distillation for compressing pretrained transformers,'' {\em arXiv preprint arXiv:2012.15828}, 2020.

\bibitem{Touvron2023LLaMAOA}
H.~Touvron, T.~Lavril, G.~Izacard, X.~Martinet, M.-A. Lachaux, T.~Lacroix, B.~Rozi{\`e}re, N.~Goyal, E.~Hambro, F.~Azhar, A.~Rodriguez, A.~Joulin, E.~Grave, and G.~Lample, ``Llama: Open and efficient foundation language models,'' {\em ArXiv}, vol.~abs/2302.13971, 2023.

\bibitem{johnson2021billion}
J.~Johnson, M.~Douze, and H.~J{\'e}gou, ``Billion-scale similarity search with gpus,'' {\em IEEE Transactions on Big Data}, vol.~7, no.~3, pp.~535--547, 2021.

\bibitem{zhang2022opt}
S.~Zhang, S.~Roller, N.~Goyal, M.~Artetxe, M.~Chen, S.~Chen, C.~Dewan, {\em et~al.}, ``Opt: Open pre-trained transformer language models,'' {\em arXiv preprint arXiv:2205.01068}, 2022.

\end{thebibliography}
}

\end{document}